# Face and Voice Cross-modal Association with Learning Convex Feature Embedding

Taewan Kim[1] and Jiwoo Kang[2*]

[1]Data Science Major, Dongduk Women's University, Seoul, 02748, South Korea.
[2*]Division of Artificial Intelligence Engineering, Sookmyung Women's University, Seoul, 04310, South Korea.

*Corresponding author(s). E-mail(s): jwkang@sookmyung.com;
Contributing authors: kimtwan21@dongduk.ac.kr;

**Abstract**

Face-and-voice association learning is one of the most challenging tasks in deep learning. In this paper, we propose a simple but powerful cross-modal feature embedding method for the association of faces and voices. Previous work has studied cross-modal association tasks to establish the correlation between voice clips and facial images. These works have addressed cross-modal discrimination but underestimate the importance of handling heterogeneity in inter-modal features between audio and video, resulting in a lot of false positives and false negatives. To tackle the problem, the proposed method learns the embeddings of cross-modal features by making another feature exist between cross-modal features, facilitating the voice and face features of the same person to be embedded in a convex hull. Moreover, the incorporation of cross-modal attention mechanisms with convex embedding techniques represents a highly effective strategy for the attenuation of false positives and false negatives, accomplished via the minimization of inter-class discrepancies. We exhaustively evaluated our method for cross-modal verification, matching, and retrieval tasks on the large-scale VoxCeleb dataset. Extensive experimental results demonstrate that the proposed method achieves notable improvements over existing state-of-the-art methods.

## 1 Introduction

Numerous earlier investigations have shown that individuals can effectively link unfamiliar faces with their corresponding voices, and vice versa [1–3]. Following the release of an extensive audio-visual dataset [4], featuring pairs of faces and voices from 1,251 celebrities, a multitude of studies have investigated the relationship between face and voice [5–15]. Tasks involving the association between faces and voices typically fall into three main categories: verification, matching, and retrieval, as illustrated in Fig. 1.

In the domain of face-voice verification, the primary objective is to determine whether pairs of facial images and vocal recordings are related to the same individual. Face-voice matching constitutes a process aimed at identifying the corresponding facial images for a specified audio clip, or conversely, the corresponding audio clip for a given facial image. Face-voice retrieval attempts to determine the facial images that exhibit the highest similarity to a given voice recording, and vice versa, by measuring the distance between the facial image vectors and voice recording vectors.

These tasks are typically performed using methodologies based on classification and metric learning [12]. In previous studies, the tasks of correlating face and voice have been approached as classification problems, where common labels are used to establish the relationship between the two modalities. Wen *et al.* [15] employ a multitask classification network to acquire knowledge about common covariates, which encompass shared attributes between the audio and video modalities, including gender, ethnicity and age. Through the training process, voices and faces are mapped into a shared feature space, allowing direct measurement of similarity between cross-modal data.

However, the performance of classification-based approaches is primarily based on the presence of shared labels. Furthermore, there are limitations in the ability to learn the semantic correlations between the face and voice data. To address this issue, recent research has employed metric learning due to its inherent flexibility. N-pair loss [16] is used in the study conducted by Horiguchi *et al.* [14]. Wang *et al.* [8] and Kim *et al.* [10] employed the triplet loss [17] to enhance the similarity between the voice and the face of people belonging to the same identity. The networks used in these studies acquire knowledge of the cross-modal similarity through supervised learning. More recently, efforts have explored unsupervised learning methods to establish voice-face associations. Nagrani *et al.* [13] employed contrastive loss [18] by incorporating a mining scheme to generate negative samples. However, this scheme is difficult to use with other metric learning methods. Recently, Chen *et al.* [5] used $k$-means clustering and multi-similarity loss [19] to learn voice-face association in an unsupervised manner. Similarly, other unsupervised methods [20, 21] use contrastive learning to establish cross-modal association using voice-face pairs from the same video clips as positive pairs while considering pairs from different clips as negative pairs. However, they simply categorize cross-modal features belonging to the same instance under the same label. It is insufficient to fully consider the differences between audio and video. The current approaches to cross-modal embedding primarily focus on cross-modal discrimination but underestimate the importance of addressing the heterogeneity in inter-modal features between audio and video. It results in a lot of false positives and false negatives, leading to limited performance improvements, as described on the left side of Fig. 2 (b).

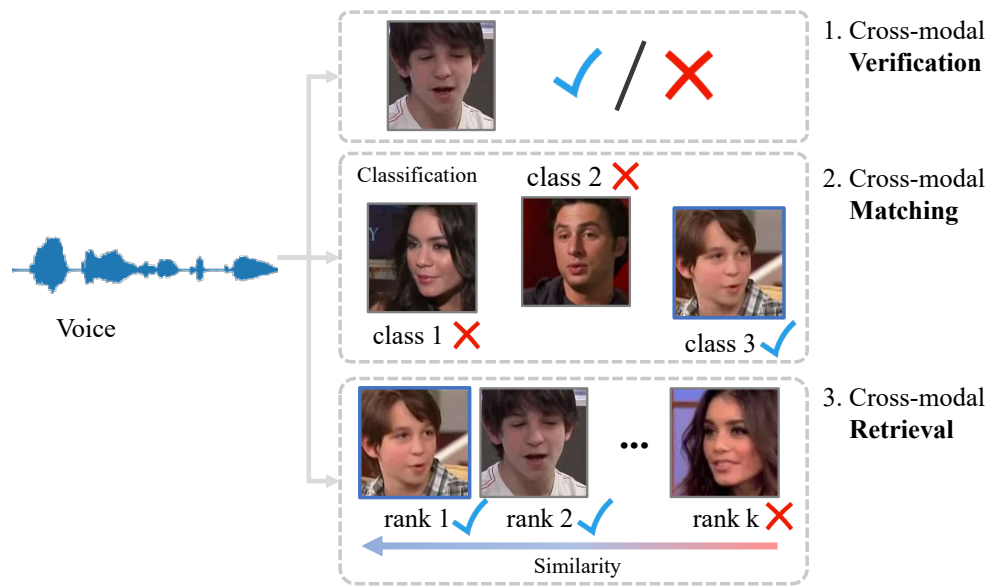

**Fig. 1**: Illustration of the three main cross-modal tasks in face-voice association: *verification*, *matching*, and *retrieval*. Verification checks whether a face and voice belong to the same person, matching identifies the correct modality from two candidates, and retrieval ranks the relevant face or voice in a larger database.

The motivation of our method stems from the large differences between the cross-modal features of recent methods. Table 1 shows the distances between face-face, voice-voice, and face-voice features, respectively. Please refer to Sect. 4.8 for more details on this evaluation. This result shows that it remains challenging to reduce the gaps between cross-modal features, although recent state-of-the-art methods [5, 7, 22] have made great improvements in the embedding of face and voice features. It is demonstrated that despite the smaller gaps between homogeneous features (*i.e.*, face-face and voice-voice distances), the significant gaps between heterogeneous features (*i.e.*, face-voice feature) make the cross-modal association limited.

To address this issue, we propose a streamlined yet robust approach for cross-modal feature embedding, specifically designed to associate faces with voice features.

Several recent studies have introduced auxiliary or intermediate modalities to bridge the semantic gap between different input sources. For instance, Wei *et al.* [23] proposed a Multi-Modality Cross Attention Network to jointly model intra-modality and inter-modality relationships for image-sentence matching. Similarly, Wei *et al.* [24] developed a universal weighting metric learning method for cross-modal matching in a shared embedding space. Peng *et al.* [25] explored sparse-to-dense feature matching in 3D semantic segmentation through intra- and inter-domain learning.

While our motivation is broadly aligned with these approaches, our method fundamentally differs in its formulation and objectives. Rather than projecting features into an abstract latent space or introducing additional branches, we explicitly model the geometric relationship between face and voice features via learnable *convex features*.

These convex features are computed as convex combinations of face and voice embeddings from the same identity, forming an identity-specific intermediary that lies within the convex hull defined by both modalities. This allows the network to maintain semantic coherence while promoting intra-class compactness and inter-class separability.

Unlike prior work, our method imposes explicit geometric constraints that ensure cross-modal embeddings of the same identity are guided toward a common convex region, while embeddings from different identities are kept distinct. To the best of our knowledge, no existing method incorporates such convex separation in the embedding space.

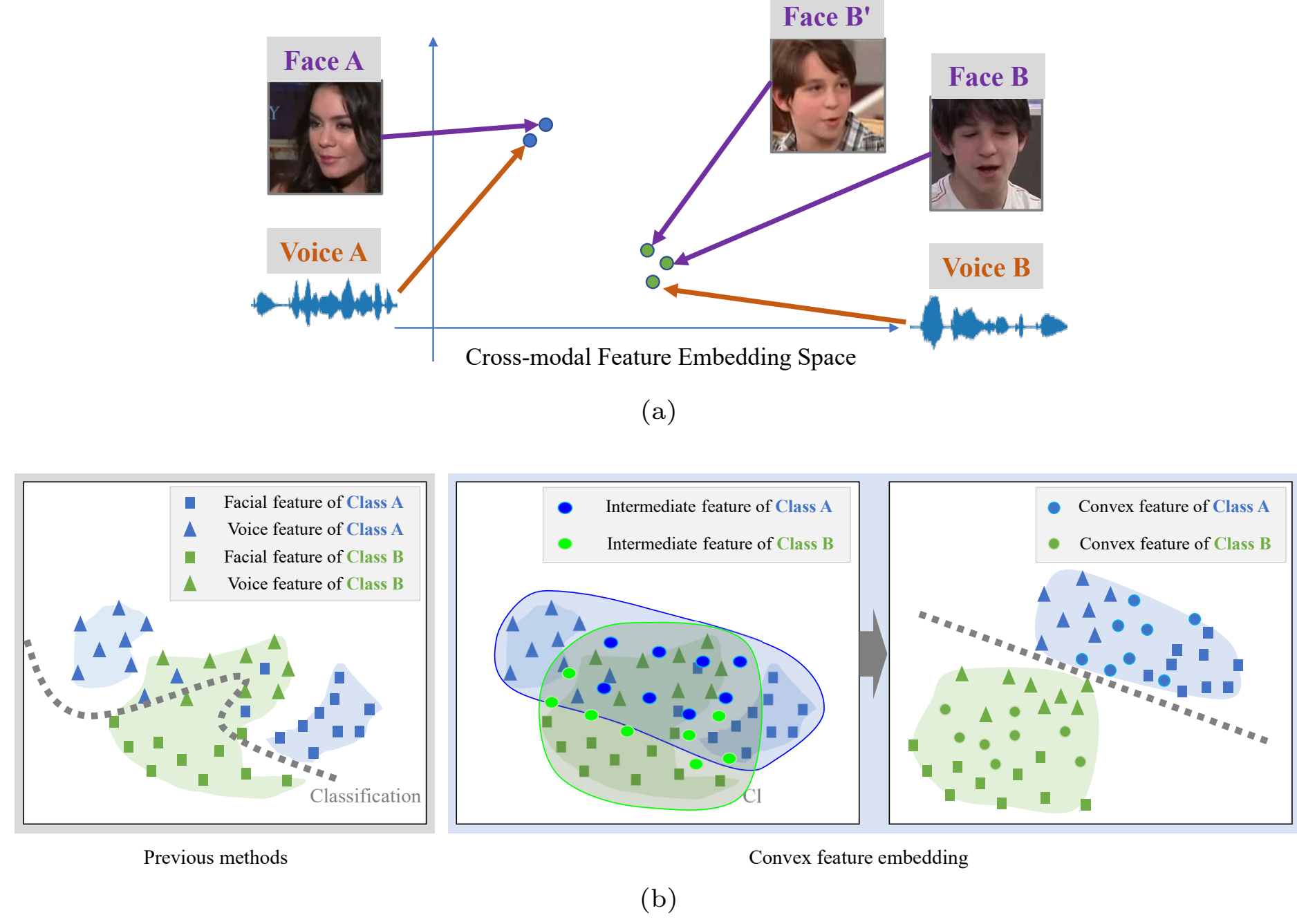

**Fig. 2**: (a) The main objective of cross-modal embedding is to efficiently align multi-modal features with identical labels in the feature space, while ensuring that features with different labels remain distinctly apart. (b) Despite this goal, the diversity of face-voice features can cause large modality gaps. Our proposed intermediate (*convex*) feature approach bridges this gap by learning a representation that lies between face and voice embeddings, thus effectively mitigating cross-modal differences.

As illustrated in Fig. 2(b), these intermediate features serve as anchors that guide face and voice embeddings toward well-separated, identity-specific regions. This strategy mitigates intra-class variation and reduces modality-induced discrepancies, ultimately leading to fewer false matches and stronger identity preservation.

Furthermore, we introduce a cross-modal attention mechanism that selectively enhances discriminative features and suppresses irrelevant signals. This mechanism complements our convex embedding by further increasing the separation between different identities, thereby reducing false positives.

The experimental results offer compelling evidence that the proposed method consistently outperforms state-of-the-art approaches on the large-scale VoxCeleb dataset [4].

In summary, this study aims to address the problem of heterogeneity in cross-modal features. To achieve this, a novel deep neural network framework is proposed, which effectively learns cross-modal feature embedding. The purpose of this framework is to facilitate face-voice association learning. The research presented in this study offers several notable contributions, which can be summarized as follows.

**Table 1**: Comparison of intra-class and inter-class cosine distances between face, voice, and cross-modal features. Lower intra-class and higher inter-class distances indicate better clustering and separation, respectively.

| Method | **(Intra-class ↓)** | | | **(Inter-class ↑)** | | |
|---|---|---|---|---|---|---|
| | Face | Voice | Cross-modal | Face | Voice | Cross-modal |
| Nagrani *et al.* [13], 2018 | 0.0224 | 0.0131 | 0.2521 | 0.1432 | 0.1378 | 0.3951 |
| Saeed *et al.* [7], 2022 | 0.0102 | 0.0085 | 0.1535 | 0.1254 | 0.1119 | 0.2680 |
| Chen *et al.* [5], 2022 | 0.0032 | 0.0055 | 0.0342 | 0.1015 | 0.0931 | 0.1476 |
| Chen *et al.* [22], 2023 | **0.0031** | 0.0049 | 0.0299 | 0.1084 | 0.0967 | 0.1592 |
| *Proposed* | 0.0035 | **0.0041** | **0.0053** | **0.1635** | **0.1510** | **0.1987** |

- The vocal and facial modalities of an individual are closely co-embedded, whereas the features of distinct individuals exhibit greater separation. This is achieved by encouraging the embedding of cross-modal features pertaining to the same individual within a convex hull.
- The integration of cross-modal attention mechanisms with convex embedding methodologies constitutes an effective approach for the mitigation of false positive and false negative instances, achieved through the reduction of inter-class gaps.
- Through comprehensive empirical analysis, our methodologies have exhibited substantial enhancements relative to current state-of-the-art frameworks in cross-modal verification, matching, and retrieval on large-scale datasets, thereby underscoring the clear benefits of the proposed approach.

This manuscript expands upon our previous investigation into face-voice association [26]. We conduct a comprehensive analysis of the implications of convex feature embedding, providing in-depth elucidation of the corresponding intricacies. In addition, we conducted a series of experiments under various conditions to rigorously validate our proposed methodology. Moreover, we perform additional comparative evaluations against contemporary approaches [5, 7, 8, 10–15, 20–22, 27].

# 2 Related Works

## 2.1 Cross-modal Learning

Multi-modal or Cross-modal learning is a widely explored area of research [28–32]. Especially, cross-modal learning of visual content and audio is a rapidly advancing domain that focuses on aligning and integrating information from these two sensory modalities to achieve deeper semantic understanding and improve downstream tasks. This field primarily concentrates on understanding the correspondence between different types of multimodal data, including text, images, and audio.

Investigating interrelationships between cross-modalities has been carried out in generation, matching, and retrieval [33–35]. Early research has primarily focused on exploring various techniques for cross-embedding images and text [36–40] Chen *et al.* [41] proposed a dual cross-modal hashing method for image-text search. Su *et al.* [42] proposed an exploration of the latent semantic relationships between images and texts. For image-to-text retrieval, Xu*et al.* [43] proposes to minimize the difference between modalities and within-class variation. Zhang *et al.* [44] proposed to use the generative adversarial network (GAN) for unsupervised manipulation of cross-modal features. Yu *et al.* [45] proposed the utilization of semantic information derived from unlabeled data. Song *et al.* [46] proposed a learning approach that uses adversarial learning to achieve joint representation. Xu *et al.* [47] employed adversarial learning techniques to address image-text retrieval problems by maximizing consistency and correlation between modalities. Chi *et al.* [48] employed two Generative Adversarial Networks (GANs) to acquire shared embedding of images and texts. Recently, many studies have been conducted on cross-modal learning in the domains of images and audio [49–52].

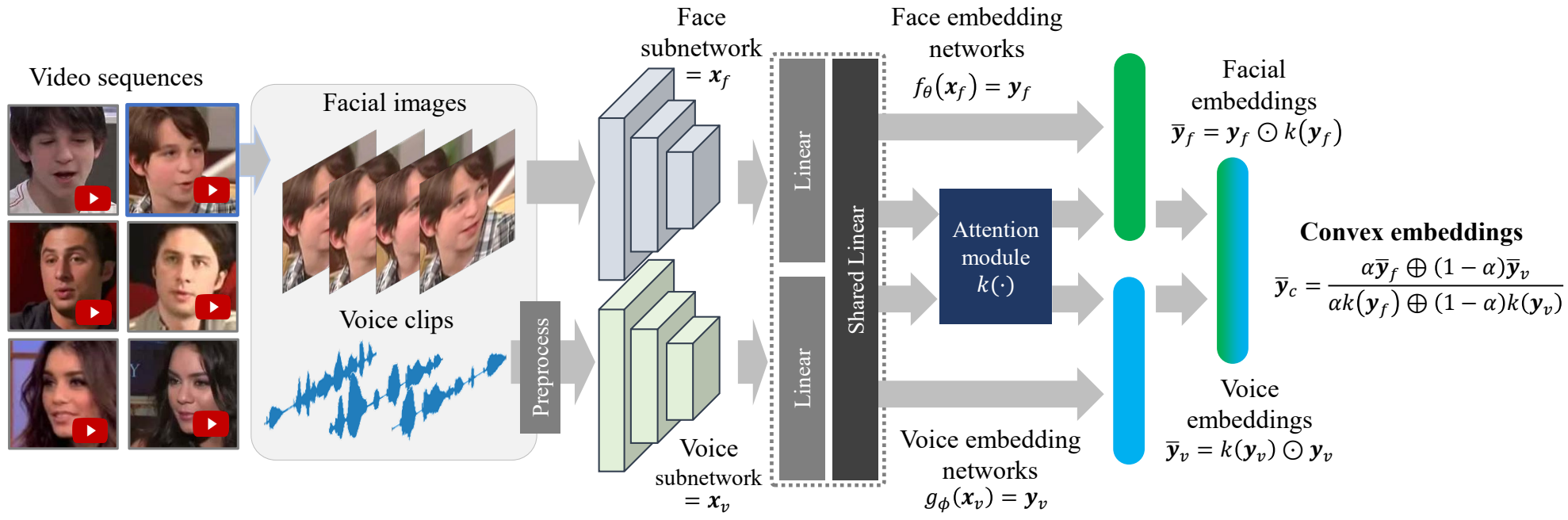

**Fig. 3**: Overview of the proposed framework. The objective of cross-modal learning is to learn the feature embedding networks, $f_\theta(\mathbf{x}_f)$ and $g_\phi(\mathbf{x}_v)$ from the face and voice features. The convex embedding facilitates the learning of precise face-voice associations within our network, as it promotes learning intermediate features that effectively bridge the gap between voices and faces. Furthermore, the incorporation of the cross-modal attention module effectively mitigates inter-class gaps by selectively attenuating disruptive signals through adaptive weighting mechanisms.

In addition, recent studies have explored more structured and high-level reasoning across modalities. For instance, Zhou *et al.* [53] proposed a structure-aware cross-modal consensus framework for video entailment, highlighting the importance of aligning abstract semantic structures across modalities. Zhang *et al.* [54] introduced a weakly-supervised audio-visual event perception framework that learns probabilistic presence-absence evidence to capture subtle yet meaningful cross-modal correlations. These approaches provide broader perspectives on multimodal understanding and complement the direction of our work.

The proposed methodology primarily emphasizes the cross-modal learning of images and audio, specifically for facial images and voices. Although other cross-modalities are beyond the scope of this paper, the convex feature embedding proposed in this paper is not restricted to cross-embedding tasks between images and audio.

## 2.2 Audio-visual Learning

Audio-visual learning includes various topics [55], such as audio-visual localization, audio-visual correspondence learning, and audio-visual generation. Audio-visual generation studies aim to generate images from sound [56], or vice versa [57], or both [58, 59]. These studies focus on recovering audio from silent videos [60] and lip movements [61, 62]. Audio-visual localization studies have localized the audio source by observing sound-image pairs [63, 64] with supervised [63, 64] and unsupervised learning [65]. In particular, the attention mechanism [66] has been shown to provide prior knowledge in semi-supervised learning by focusing on the primary area. Inspired by these works, we introduce an attention network to decrease gaps between intra-class features and encourage the clustering of cross-modal features. Studies of audio-visual correspondence learning have investigated the semantic relation between the audio-visual modalities. These works focus mainly on the audio-visual recognition task and the audio-visual matching task.

In this paper, we propose a novel learning method to better learn the audio and visual modalities, especially for face-and-voice pairs. Our method is evaluated in various audio-video tasks, including verification, matching, and retrieval tasks.

## 2.3 Face-voice Association Learning

In recent years, many efforts have been made to explore the relationship between the face and voice of individuals [5, 10, 13, 27, 67, 68] since the first work on voice-face matching and retrieval tasks [69]. These methods are designed to enhance the separation between the audio and facial features of individuals belonging to different identities while simultaneously reducing the distance between the features of the same identity. Zheng *et al.* [70] combined metric and adversarial learning to generate a modality-independent representation. Wen *et al.* [15] introduced a disjoint mapping network to acquire a shared representation for audio and visual modalities using common covariates between cross-modal information, such as identity. Nagrani *et al.* [13] employed a self-supervised approach to acquire cross-modal embeddings by utilizing contrastive loss [18]. Horiguchi *et al.* [14] employed the N-pair loss [16] to enhance the correlation between the face and the voice of the same identity. Similarly, Kim *et al.* [10] the triplet loss [17] to establish the association of face and voice. More recently, Chen *et al.* [5] used the $k$-means multi-similarity loss [19] to learn the association of voice and face without supervision. Besides, Morgado *et al.* [20] and Zhu *et al.* [21] employed contrastive learning to acquire cross-modal associations using voice-face pairs obtained from video clips.

In contrast to the aforementioned approaches for associating faces with voices, our method primarily addresses the inherent heterogeneity between audio and video data. Our convex feature approach effectively addresses the challenges, leading to notable improvements compared to existing state-of-the-art methods.

# 3 Cross-modal Embedding Learning

The proposed architecture to embed convex features for cross-modal analysis is described in Fig. 3. Let $\mathbf{x}_f$ and $\mathbf{x}_v$ be features extracted from faces and voices using dedicated subnetworks for face and voice processing. Numerous studies have been conducted to extract features from facial images [71–73] and voice clips [74–76] using deep neural networks. In line with this, our framework leverages these established models for the face and voice subnetworks.

Our goal is to learn the *feature embedding network*s, $f_\theta(\mathbf{x}_f)$ and $g_\phi(\mathbf{x}_v)$ from the facial and voice features. The *feature embedding network*s learn common embedding to jointly represent face and voice. In other words, face and voice feature embedding of individuals belonging to the same identity become closer together, while those of different identities become distant away.

## 3.1 Framework

Our convex feature embedding method can be applied in supervised or unsupervised manners. Without losing generality, we present our methodology in an unsupervised fashion. Thus, we follow a recent work on face-voice association [5] for the purpose of unsupervised learning. We use video sequences with a single speaker [4], which is consistent with numerous prior approaches to face-voice association [5, 13, 20, 21]. Multiple facial images and voice clips are extracted from a single video sequence. If the pairs of facial images and the corresponding audio recordings are derived from the same video sequence, they are categorized as positive pairs. Similarly, negative pairs can be defined as combinations of faces and voices from different videos. However, given that distinct video sequences may possess identical identities, it becomes necessary to reassign labels to video sequences that share the same identity.

Supervised learning methods [7, 14, 15, 27] operate under the assumption that identity labels are available, allowing the categorization of sequences according to their ground-truth labels. However, in the context of unsupervised learning methods [5, 13, 20, 21], where identity labels are assumed to be unknown, it becomes necessary to employ mining or clustering techniques to categorize similar features prior to learning cross-modal feature embedding. To categorize similar features, the techniques of k-means clustering and multi-similarity loss [19] are employed to learn the association of the voice and the face in an unsupervised manner, which is consistent with recent research [5]. Here, cosine similarity is used to measure the distance between cross-modal feature embeddings, $\mathbf{y}_f$ and $\mathbf{y}_v$, for $k$-means clustering. Subsequently, the results obtained from the clustering are utilized to learn the similarity between the features through the metric learning process. Thus, the *feature embedding network*s, denoted as $f_\theta(\mathbf{x}_f)$ and $g_\phi(\mathbf{x}_v)$, are trained using metric learning techniques. The utilization of the clustering method to train the *feature embedding network* mainly relies on the labeled face and voice feature embedding. On the contrary, the accuracy of the clustering process is significantly influenced by the performance of the *feature embedding network*. Therefore, the clustering and metric learning steps are performed iteratively until the *feature embedding network* converges.

We propose simple but robust convex embedding in order to facilitate the network's acquisition of precise face-voice associations for clustering and metric learning. This objective is achieved by promoting the learning of intermediate features that bridge the gap between voices and faces. Furthermore, we propose the incorporation of a cross-modal attention module, denoted as $k(\cdot)$, which allows for weighting face and voice embedding through attention scores. The attention mechanism plays a crucial role in filtering out irrelevant signals, thereby minimizing the variability among different classes and leading to a significant reduction in both false positives and false negatives. In the subsequent section, we will begin by introducing the framework architecture. This will be followed by a description of the loss function that is utilized within our framework.

## 3.2 Architecture and Methodology

### 3.2.1 Face and Voice Subnetworks

Face and voice features, $\mathbf{x}_f \in \mathbb{R}^{n_f}$ and $\mathbf{x}_v \in \mathbb{R}^{n_v}$, are obtained, respectively, from RGB images and voice clips by pre-trained networks [72, 74].

Specifically, we adopt Inception-V1 [72] (pre-trained on VGGFace2 [71]) for the face subnetwork. Each input face image is resized to $224 \times 224$, then passed through the truncated Inception-V1 up to the final global average pooling layer, resulting in a 512-dimensional face feature vector. For the voice subnetwork, we use ECAPA-TDNN [74, 75], which produces a 192-dimensional voice feature vector from an audio clip sampled at 16 kHz. We do not fine-tune these pretrained models; rather, we freeze their parameters and only train subsequent layers, as we found that end-to-end fine-tuning led to overfitting on the relatively small dataset. This approach also reduces computational cost while leveraging the robust representations learned by the pretrained networks.

### 3.2.2 Feature Embedding Networks

The *feature embedding network*s, $f_\theta(\cdot)$ and $g_\phi(\cdot)$. learn to generate *facial embedding*s and *voice embedding*s, $\mathbf{y}_f \in \mathbb{R}^{n_e}$ and $\mathbf{y}_v \in \mathbb{R}^{n_e}$, respectively, from facial and voice features, $\mathbf{x}_f$ and $\mathbf{x}_v$. Specifically, a fully connected layer is used separately for each modality to project face and voice features into the $n_m$ dimensional intermediate (*i.e.* hidden) embedding space. Subsequently, inspired by previous work [8], we used two shared fully-connected layers for both modalities after the hidden embeddings, producing the feature embeddings, $\mathbf{y}_f$ and $\mathbf{y}_v$.

In our implementation, we set $n_m = 256$ and $n_e = 256$, such that both face and voice embeddings are mapped into the same 256-dimensional space. Each fully connected layer is followed by a ReLU activation, and we optionally apply dropout ($p = 0.2$) to mitigate overfitting. This shared embedding space facilitates direct comparison between facial and vocal features, as subsequent components (*e.g.*, cross-modal attention, convex feature computation) rely on both embeddings having consistent dimensionality. We found that using separate embedding layers for face and voice, then merging them with two shared layers, strikes a balance between modality-specific processing and cross-modal alignment.

### 3.2.3 Cross-modal Attention

We implement the cross-modal attention network $k(\cdot)$ as a lightweight multi-layer perceptron (MLP) that outputs attention scores for each dimension of the embedding. Concretely, let $\mathbf{y} \in \{\mathbf{y}_f, \mathbf{y}_v\}$ be an input embedding of dimension $n_e$. The attention network applies two consecutive fully-connected layers, each with $n_m$ hidden units, followed by a ReLU activation. The final layer projects back to $n_e$ dimensions, and a sigmoid function $\sigma(\cdot)$ normalizes each component to the range $[0, 1]$. Formally,

$$k(\mathbf{y}) = \sigma\big(W_2 \, \mathrm{ReLU}(W_1 \, \mathbf{y} + \mathbf{b}_1) + \mathbf{b}_2\big), \tag{1}$$

where $W_1, W_2, \mathbf{b}_1, \mathbf{b}_2$ are learnable parameters. The resulting vector $k(\mathbf{y}) \in [0, 1]^{n_e}$ represents per-dimension attention scores that emphasize important features while suppressing irrelevant ones. Since we aim to learn a unified cross-modal attention, this network is shared by both face and voice embeddings, ensuring modality-agnostic processing.

Therefore, *facial embedding*s and *voice embedding*s weighted by the attention scores, $\bar{\mathbf{y}}_f$ and $\bar{\mathbf{y}}_v$, are obtained, respectively, as

$$\begin{aligned} \bar{\mathbf{y}}_f &= \mathbf{y}_f \odot k(\mathbf{y}_f), \\ \bar{\mathbf{y}}_v &= \mathbf{y}_v \odot k(\mathbf{y}_v), \end{aligned} \tag{2}$$

where $\odot$ denotes the operator of element-wise multiplication.

In essence, each dimension of $\mathbf{y}_f$ (or $\mathbf{y}_v$) is modulated by the corresponding attention score in $k(\mathbf{y}_f)$ (or $k(\mathbf{y}_v)$). High attention values amplify salient features, whereas lower values attenuate noisy or less discriminative components. This mechanism ensures that the subsequent convex feature computation leverages the most informative aspects of each modality.

### 3.2.4 Convex Feature

The ultimate goal of the voice-face association is to learn common embeddings to represent both face and voice that is indistinguishable. However, the feature embeddings obtained from faces and voices differ significantly due to their inhomogeneity. To efficiently handle cross-modal differences during network training, we introduce the *convex feature embedding*s, *i.e.* intermediate feature embeddings between cross-modal embeddings. Thus, for an arbitrary value $\alpha \in [0, 1]$, the *convex feature embedding* is defined as

$$\bar{\mathbf{y}}_c = \frac{\alpha \bar{\mathbf{y}}_f \oplus (1 - \alpha) \bar{\mathbf{y}}_v}{\alpha k(\mathbf{y}_f) \oplus (1 - \alpha) k(\mathbf{y}_v)} \tag{3}$$

where $\oplus$ is the element-wise addition operator.

We sample $\alpha$ from a uniform distribution over $[0, 1]$ for each instance, which serves as the default (baseline) configuration in our framework.

This stochastic sampling serves as the default (baseline) configuration in our framework. To examine the effect of this parameter, we also conduct an ablation test using a fixed value of $\alpha = 0.5$, as shown in Table 2 of the experimental section.

Although the attention scores appear in the denominator of Eq. (3), this formulation does not reduce the influence of highly attended features. Instead, the denominator acts as a normalization factor, scaling the convex combination such that modalities with higher attention scores contribute proportionally more to the final convex feature. This ensures that the convex feature embedding reflects a weighted interpolation that respects the relative importance of each modality, rather than suppressing high-attention components.

Here, both the face/voice embeddings $(\bar{\mathbf{y}}_f, \bar{\mathbf{y}}_v)$ and their respective attention scores $(k(\mathbf{y}_f),\ k(\mathbf{y}_v))$ are combined via weighted interpolation. While it may appear counterintuitive that dividing by the sum of attention scores could diminish the influence of a modality with high attention, this formulation actually creates a *normalized weighted average* in which higher attention yields a proportionally larger share of the final embedding. In other words, a modality with higher $k(\mathbf{y})$ also has a greater contribution in the numerator, thus ensuring that its overall impact is amplified rather than reduced. This normalization is crucial for keeping the resulting vector within a consistent range and preserving the convex hull property.

This formulation can be interpreted as a normalized convex combination of attention-weighted face and voice embeddings. The numerator computes a linear interpolation of the face and voice embeddings weighted by $\alpha$, while the denominator computes a corresponding interpolation of their attention scores. This normalization ensures that the resulting convex feature lies within a scaled convex hull of the two embeddings, with adaptive weighting based on the attention responses.

The random sampling of $\alpha$ at each iteration helps the network explore a variety of interpolated representations between the two modalities, thereby enriching the learned cross-modal structure. Since $k(\cdot)$ produces values in $[0, 1]$, the denominator serves to scale the interpolated embedding according to the overall importance of each modality.

Overall, this convex embedding encourages the model to project intermediate representations closer to the intra-class cluster center, while still respecting the modality-specific contribution indicated by the attention. This strategy helps to close the modality gap and improve the discriminability of cross-modal features. The loss for convex feature embedding helps a deep neural network to learn the cross-modal feature embeddings in a convex hull, encouraging inter-modal features and inter-class features to be projected into the closer space.

### 3.2.5 Clustering and Metric Learning

The convex features, $\bar{\mathbf{y}}_c$ obtained from pairs of faces and voices of video sequences, are used for the training set for $k$-means clustering ($k = 1000$). Furthermore, the triplets of feature embeddings, which are *facial*, *voice*, and *convex feature embedding*s, are used to train the network with metric learning. Thus, we use the batch size that is a multiple of 3. For metric learning, we use the multi-similarity loss [19].

During each iteration, we assign pseudo-labels to the embeddings through $k$-means clustering, and then update the network parameters by minimizing the multi-similarity loss over these clustered triplets. In practice, each triplet consists of an anchor sample, a positive sample from the same cluster, and a negative sample from a different cluster. Because multi-similarity loss considers all possible positive and negative pairs within a batch, the network effectively learns relative distances in a unified embedding space. This iterative process continues until convergence, allowing the network to progressively refine both the clustering quality and the cross-modal embedding itself.

# 4 Experimental Results

## 4.1 Experimental Details

### 4.1.1 Dataset

We performed our experiments on cross-modal verification, matching, and retrieval tasks on VoxCeleb [4], which is the large-scale face-voice dataset. The dataset comprises 21,063 video clips from 1,251 celebrities extracted from YouTube videos and has been used in many previous works [5–7, 9, 13, 14, 22]. We divided the dataset into train, test, and validation splits according to the configuration of recent works [5, 22] for cross-modal verification, matching, and retrieval tasks. Note that there are no overlapping identities between the training, test, and validation splits. The test set is annotated with three different demographic criteria: gender (G), nationality (N), and age (A). Furthermore, the combination of criteria (*i.e.* GNA) and the unconstraint group (U) are used to examine the effects of the criteria.

### 4.1.2 Face and Voice Feature Sub-networks

Previous works [5, 8, 13, 14] reported that feature extraction architectures for facial images and voice clips affect overall performance. Therefore, we follow the architecture used in recent work on the face-voice association [5] for fair comparisons. Specifically, *Inception-V1* [72] pre-trained in *VGGFace2* [71] is used for extraction of facial features. *ECAPA-TDNN* [74, 75] pre-trained on *VoxCeleb2* [77] is used for the extraction of voice features. The feature vectors extracted from a facial image and a voice clip have 512 and 128 dimensions, respectively ($n_f = 512$, $n_v = 128$). Also, we use 256-dimensional features for every intermediate fully-connected layer in our experiments ($n_m = 256$). As we do not update feature extraction networks, these vectors are extracted and saved before training to maximize efficiency. For fair comparisons, we used common networks to extract face and video features from video clips, respectively, for all methods in our experiments.

### 4.1.3 Implementation Details

Our networks are trained on a single *NVIDIA GeForce RTX 3090* GPU for 50 *epochs* with a batch size of 384. We implement our method with a deep learning framework, *PyTorch* [78]. The multi-similarity loss we used for unsupervised metric learning is based on the *Pytorch Metric Learning* library [79]. We use the *Adam* optimizer [80] with an initial learning rate $10^{-3}$ with an exponential decay rate of 0.9.

**Table 2**: Verification results with ablation tests. We systematically remove or modify components (*e.g.*, convex embedding, attention layer,hidden size, and shared layers) to evaluate their impact on performance. The parameter $\alpha$ is also tested with a fixed value (0.5) versus random sampling.

| Config | Method | Verification AUC % (↑) | Verification EER % (↓) |
|---|---|---|---|
| Baseline | Proposed | **89.71** | **19.19** |
| Network | w/o convex feature | 86.97 | 21.71 |
| | w/o attention layer | 87.51 | 20.95 |
| Hidden Size | 128 | 88.97 | 20.01 |
| | 256 | 89.71 | 19.19 |
| | 512 | 89.23 | 19.51 |
| Shared FC | Not Shared | 88.12 | 21.11 |
| Attention | Not Shared | 87.98 | 20.74 |
| Parameter $\alpha$ | Constant (0.5) | 88.50 | 20.51 |
| Loss | Contrastive loss | 85.12 | 22.55 |
| | Triplet loss | 86.42 | 21.82 |
| | Lifted Struct loss | 86.12 | 21.90 |
| | Center loss | 86.53 | 21.74 |
| | N-pair Struct loss | 86.91 | 21.71 |

#### 4.1.4 Evaluation Metrics

We evaluate our method on three cross-modal subtasks: *verification*, *matching*, and *retrieval*, following standard practices in cross-modal research [5–7, 9].

**Verification.** For the face-voice verification task, we report the Area Under the ROC Curve ($AUC$) and Equal Error Rate ($EER$) [81]. AUC reflects the ability of the model to distinguish between same-identity and different-identity pairs, while EER is the point where false acceptance and false rejection rates are equal; hence, lower EER indicates better performance.

**Matching.** For cross-modal matching (1:2 matching), we report *accuracy* ($ACC$) [69], which measures whether the correct modality is selected from two candidates. Matching tasks are conducted in both directions: voice-to-face (V-F) and face-to-voice (F-V), with additional evaluations under gender constraints.

**Retrieval.** For retrieval, we adopt *mean Average Precision* ($mAP$) [82], which evaluates the quality of ranked retrieval results based on the average precision across queries. mAP is especially useful in evaluating retrieval tasks where multiple correct results may exist in the ranked list. Higher mAP values indicate more relevant and accurate retrieval performance.

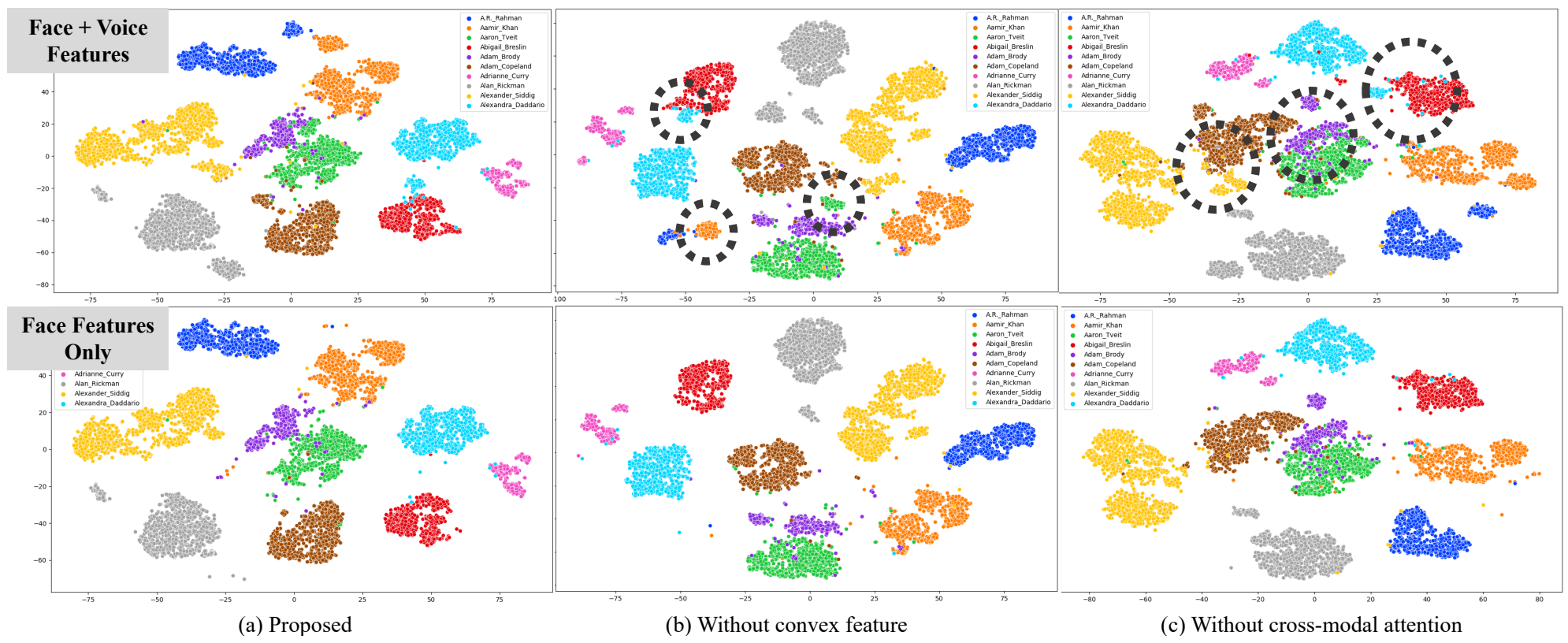

(a) Proposed (b) Without convex feature (c) Without cross-modal attention

**Fig. 4**: T-SNE visualization of voice and face representations. The first ten subjects in the test set are chosen. Top: both face and voice features. Bottom: face features only. The result shows that convex feature embedding helps gather cross-modal features from the same person, while embeddings from different people are clearly separated. In addition, cross-modal attention prevents overlap between features from different classes.

## 4.2 Ablation Study

We conducted several experiments with different configurations to fully evaluate our method through the ablation study. First, we compared the results with those without convex feature embedding and cross-modal attention to examine the role of each component. Second, we evaluated our method by differing the sizes of the intermediate layer (*i.e.* $n_m$ =128, 256, or 512). Third, we used separate networks for face and voice embedding, respectively, in the *feature embedding network*s and the *cross-modal attention network* to validate the benefits of using a common network for different modalities.

Fourth, we examined the effect of the convex interpolation parameter $\alpha$ in (3). By default, $\alpha$ is randomly sampled from the interval $[0, 1]$ for each instance, which serves as the baseline setting in our method. To evaluate its contribution, we compared this against a fixed constant setting of $\alpha = 0.5$.

The results of the verification task obtained by varying configurations are summarized in Table 2 in terms of AUC and equal error rate (EER). The convex feature embedding considerably improves performance in terms of AUC and EER, indicating that it contributes mainly to the performance gain of our cross-modal association method. In addition, the attention layer is shown to help drive more accurate cross-modal association tasks in conjunction with convex feature embedding. The ablation study demonstrates that the use of both convex feature embedding and cross-modal attention leads to more accurate cross-modal embeddings.

Furthermore, using fewer features for intermediate layers than 256 distinctly decreases the performance. On the contrary, results that use more features for intermediate layers result in comparable results, but this is not beneficial with respect to performance and memory requirements.

The shared structures in *feature embedding network*s and *cross-modal attention network* are shown to help accurately learn joint embeddings as they encourage undistinguishable cross-modal features. The shared network helps to correlate heterogeneous facial and voice features. In particular, the result demonstrates that a common network is notably beneficial for the *cross-modal attention network*.

Regarding the parameter $\alpha$, the results show that randomly sampling $\alpha$ during training leads to better performance than using a fixed value. This stochastic interpolation allows the model to explore diverse intermediate embeddings between modalities, thereby improving generalization and reducing overfitting to a specific modal balance. The performance drop with $\alpha = 0.5$ demonstrates the advantage of preserving this randomness during training.

We compared multi-similarity loss [19] with contrastive loss [18], triplet loss [17], lifted struct loss [83], center loss [84], and N-pair struct loss [16]. Among the metric learning losses, MS loss shows the best performance, since MS loss considers all features in a batch to measure each distance. In contrast, the contrastive loss shows poor results in learning cross-modality as it simply considers the positive and negative relationship of each pair. The result shows that MS loss is more appropriate than other metric learning losses for learning multi-modal features. Thus, by integrating MS loss, the proposed method accurately learns cross-modal embeddings.

### 4.3 Qualitative Analysis

We visualize the learned embedding features of face and voice features using *t-SNE* method [85]. We choose the first ten people in the test set and visualize the facial and voice embedding features, as depicted in Fig. 4.

To analyze the interaction between modalities, we present each case in two rows: the first row shows *both face and voice embeddings* jointly, and the second row shows *only face embeddings*. This design highlights whether the embeddings from different modalities (face and voice) of the same identity are well-aligned in the feature space. By comparing the two rows, we can examine how voice features contribute to modality alignment and whether the model achieves compact clusters in the joint (cross-modal) embedding space.

The result shows that the convex feature embedding helps to gather the face and voice embeddings of the same person, whereas the embeddings of different people are clearly apart. In particular, the convex feature embedding helps the features embedded in the convex hull, whose boundaries are constructed by features in the training set. Without the convex feature embedding, the cross-modal features with different modalities are embedded in different spaces, as described in Fig. 4 (b). The result demonstrates that the convex feature embedding significantly decreases false positives and efficiently addresses cross-modal inhomogeneity.

Moreover, Fig. 4 (a) shows that under the proposed method, facial and vocal embeddings from the same identity are projected into overlapping clusters, indicating that voice features effectively contribute to intra-class compactness across modalities. In contrast, Fig. 4 (b) and (c) reveal that removing either convex features or cross-modal attention results in misaligned embeddings, where face and voice features from the same identity fail to form coherent clusters. The dotted circles in (b) and (c) highlight voice embeddings that are not properly aligned with their corresponding face features, indicating that the model struggles to bridge the modality gap without the proposed components.

Cross-modal attention is shown to prevent overlap between features of different classes, as described in Fig. 4 (c). The result demonstrates that cross-modal attention helps to make distinct boundaries between different classes by decreasing inter-class gaps. The convex feature embedding and the cross-modal attention have distinct benefits in learning cross-modal features, and using both is the most helpful to accurately handle differences between multi-modal features, as described in Fig. 4 (a). These results are consistent with those of Table 2, demonstrating the effectiveness of our method in handling inhomogeneity of cross-modal features.

## 4.4 Comparisons with State-of-the-art Methods

### 4.4.1 Methods

Several recent supervised [7, 8, 10–12, 14, 15, 22, 27] and unsupervised methods [5, 13, 20, 21] for cross-modal association are used to compare the performance with our method in cross-modal matching, verification, and retrieval.

### 4.4.2 Results

We evaluated our network on the *cross-modal verification* task. The goal is to determine whether the face-and-voice pair belongs to the same identity. All verification results were performed on test datasets, which are *unseen-unheard* data in the training stage. We also evaluated our network on the *cross-modal matching* and *retrieval* tasks. In the face-to-voice (F-V) matching task, for a given facial image and two audio clips, the audio clip that corresponds to the facial image is found, and vice versa in the voice-to-face (V-F) matching task. The retrieval task is an extension of the cross-modal matching task. First, ten audio clips and ten facial images are selected from each person. Then, in the F-V retrieval, every item in the ten faces is used to retrieve from the ten voices, and vice versa in the V-F retrieval. We used the best model on the validation set to measure the results on the test set. The results are summarized in Table 3.

The gender constraint setting, *i.e.* testing with the same gender, significantly decreases the overall performance across all the methods. This implies that there are large differences in voice and facial features between different genders. Nevertheless, the performance order does not change despite the overall performance reduction. The proposed method shows significant improvements over existing methods in cross-modal verification, matching, and retrieval. In particular, the proposed methods achieved nearly 90% performance in the verification task.

**Table 3**: Comparison with other supervised (*Sup.*) and unsupervised (*Unsup.*) Methods in verification task, 1:2 matching task, and retrieval task using unified feature extraction subnetworks. The $U$ and $G$ stand for unconstraint setting and the same-gender constraint setting, respectively. Training our network in an unsupervised manner is shown to be more beneficial than supervised training.

| Type | Method | Verification (AUC %) | | 1:2 Matching (ACC) | | | | Retrieval (mAP) | |
|---|---|---|---|---|---|---|---|---|---|
| | | U | G | V2F (U) | V-F (G) | F-V (U) | F-V (G) | V-F | F-V |
| Sup. | Wen *et al.* [15], 2018 | 84.69 | 73.84 | 84.02 | 73.00 | 84.23 | 73.26 | 6.22 | 6.65 |
| | Xiong *et al.* [12], 2019 | 84.77 | 71.09 | 83.46 | 70.82 | 83.91 | 71.34 | 4.70 | 5.47 |
| | Horiguchi *et al.* [14], 2019 | 85.48 | 71.48 | 84.23 | 72.09 | 84.69 | 72.44 | 5.06 | 5.79 |
| | Nawaz *et al.* [11], 2019 | 86.35 | 75.45 | 85.48 | 74.13 | 85.78 | 74.42 | 6.87 | 7.57 |
| | Kim *et al.* [10], 2019 | 86.65 | 76.01 | 85.65 | 74.60 | 86.17 | 74.76 | 6.74 | 7.56 |
| | Wang *et al.* [8], 2020 | 86.99 | 76.70 | 86.07 | 74.96 | 86.59 | 75.21 | 6.78 | 7.73 |
| | Saeed *et al.* [7], 2022 | 85.40 | 75.97 | 83.33 | 73.89 | 84.56 | 74.15 | 5.95 | 7.48 |
| | Chen *et al.* [22], 2023 | 87.10 | 77.13 | 86.21 | 75.82 | 86.52 | 75.19 | 7.22 | 7.95 |
| | *Proposed (Supervised)* | **89.40** | **80.20** | **88.34** | **78.61** | **88.19** | **77.75** | **9.03** | **9.39** |
| Unsup. | Nagrani *et al.* [13], 2018 | 81.88 | 63.09 | 80.82 | 65.92 | 80.98 | 65.94 | 3.34 | 3.58 |
| | Chen *et al.* [5], 2022 | 86.70 | 76.96 | 85.86 | 75.07 | 85.92 | 74.73 | 6.94 | 7.65 |
| | *Proposed (Unsupervised)* | **89.71** | **80.85** | **88.79** | **78.69** | **88.63** | **78.32** | **9.21** | **9.47** |

### 4.4.3 Supervised versus Unsupervised Learning

Table 3 shows several methods trained in supervised or unsupervised manners. Here, we also measured our results with supervision, where ground-truth labels were used to determine clusters for faces and voices instead of using $k$-means clustering. The results reported in the unsupervised work of Chen *et al.* [5] and Zhu *et al.* [21] are comparable to the supervised work of Saeed *et al.* [7] and Chen *et al.* [22]. Notably, our supervised approach does not yield significant improvements over the unsupervised one; in fact, performance is slightly degraded in some cases.

This observation can be theoretically understood through three well-established principles in representation learning. First, based on the *manifold hypothesis*, real-world data such as facial images and voice clips lie on a low-dimensional manifold [86]. Clustering-based methods exploit this by allowing the model to learn the intrinsic structure of the data without being constrained by hard categorical labels. This enables the embedding function to discover more flexible, semantically meaningful partitions and better align heterogeneous modalities.

Second, supervised approaches may suffer from overfitting to training identity labels, especially when label distributions are imbalanced or limited in variability. In contrast, unsupervised metric learning relies on relative similarities, which inherently support better generalization across unseen identities [87]. This aligns with empirical results where test-time generalization benefits from avoiding rigid class assignments.

Third, our approach benefits from principles similar to *Noise Contrastive Estimation (NCE)* and the *InfoMax* principle [88], which emphasize maximizing mutual information between associated modalities. By constructing convex combinations of face and voice embeddings, our method effectively increases modality-invariant representation capacity, helping the model focus on shared identity cues and ignore modality-specific noise. These theoretical foundations offer a principled explanation for the strong generalization observed in our unsupervised results.

This perspective also explains why our unsupervised framework performs particularly well in matching and retrieval tasks, which are inherently driven by distance-based similarity rather than absolute class assignments. Since our method optimizes feature distances through iterative clustering and metric learning, it aligns naturally with retrieval-style evaluation protocols. Although no ground-truth labels are used during training, the clustering process effectively generates identity-consistent pseudo-labels, allowing the model to learn semantically meaningful embeddings even without explicit supervision.

## 4.5 Evaluation on Cross-modal Gap

To further analyze the benefits of our method in addressing the heterogeneous issue, we evaluated the Euclidian distance between the cross-modal face and voice features with and without the convex feature embedding and the cross-modal attention. In particular, we measured the average distance between face and voice features of the same identity for three cross-modal tasks: verification, matching, and retrieval tasks.

Fig. 5 visualizes the results. It is shown that both the convex feature embedding and the cross-modal attention help to significantly decrease the cross-modal differences.

**Table 4**: Verification with other feature subnetworks. Although overall performance degrades due to different backbones, the proposed method still outperforms existing approaches, indicating its robustness to changes in feature extraction networks.

| Method | Verification (U) | |
|---|---|---|
| | AUC % (↑) | EER % (↓) |
| Nagrani *et al.* [13], 2018 | 80.2 | 27.8 |
| Saeed *et al.* [7], 2022 | 83.5 | 24.9 |
| Chen *et al.* [5], 2022 | 84.1 | 24.5 |
| Chen *et al.* [22], 2023 | 84.9 | 23.5 |
| *Proposed* | **87.9** | **20.6** |

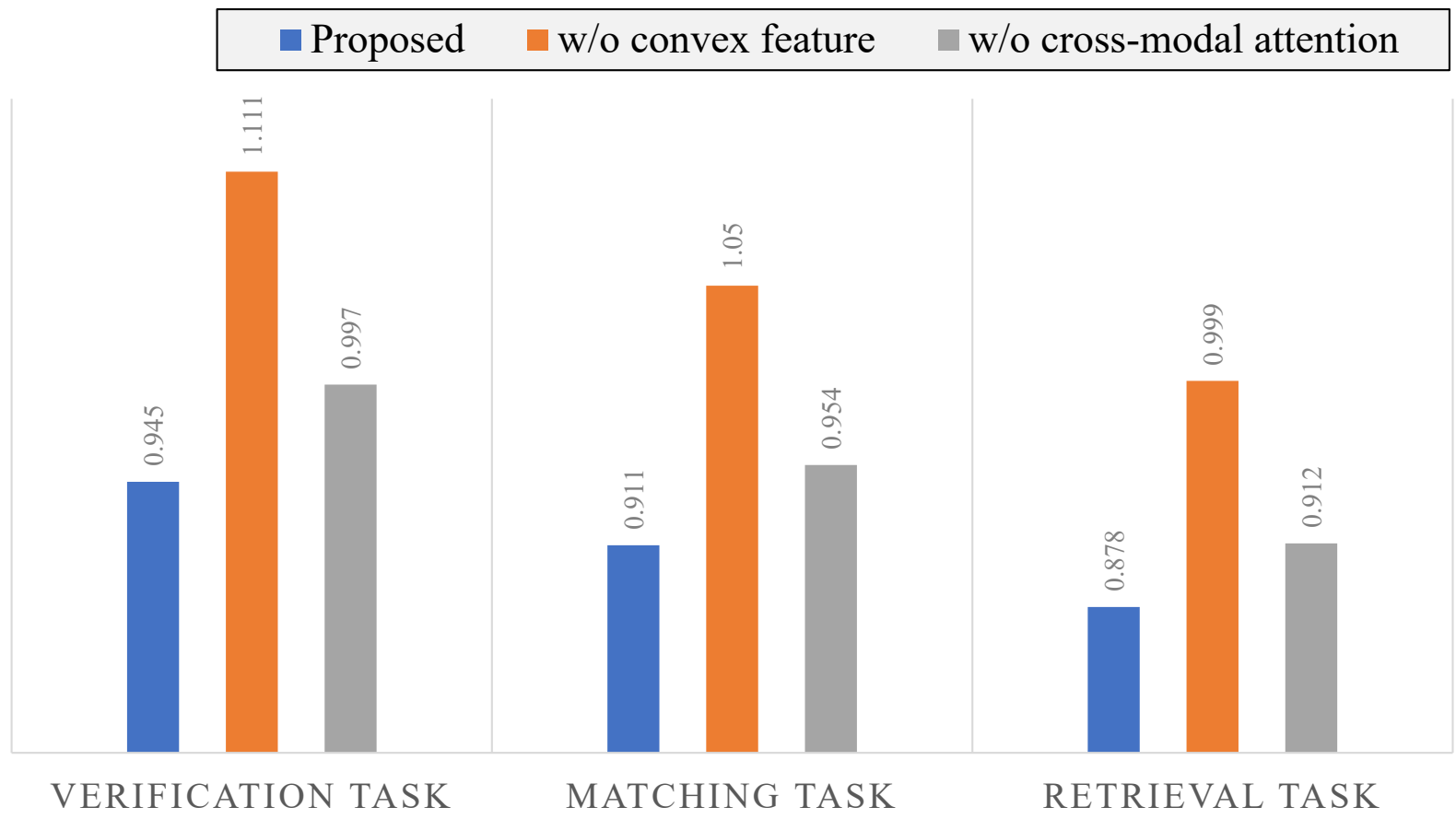

**Fig. 5**: Average Euclidean distance between face and voice features. Lower distance indicates better alignment of cross-modal embeddings. It is shown that the convex feature embedding and the cross-modal attention significantly lower the cross-modal differences, demonstrating that our method efficiently addresses heterogeneity for cross-model embedding.

The results demonstrate the effectiveness of our method in addressing heterogeneity for learning cross-model embedding.

## 4.6 Evaluation on Other Feature Embeddings

To validate our method in different feature subnetworks, we evaluated the performance using other network architectures to extract facial and voice features. For evaluation, we used subnetworks used in [7]. In detail, we extracted face and voice feature embeddings, respectively, from *VGGFace* [73] and Utterance Level Aggregation [76] pre-trained on *VoxCeleb1* [4] and *VoxCeleb2* [77].

**Table 5**: Cross dataset split validations and comparisons with state-of-the-art methods on verification task, 1:2 matching task, retrieval task. The results are consistent with the results in the other split validations, demonstrating the stability and effectiveness of our approach across different experimental setups.

| Type | Method | Verification (AUC %) | | | | | V-F Matching (ACC %) | | | | F-V Matching (ACC %) | | | | Retrieval (mAP %) | |
|---|---|---|---|---|---|---|---|---|---|---|---|---|---|---|---|---|
| | | U | G | N | A | GNA | U | G | N | GN | U | F | N | GN | V-F | F-V |
| Sup. | Wen *et al.* [27], 2018 | 83.2 | 71.2 | 81.9 | 78.0 | 62.8 | 83.5 | 70.9 | 82.0 | 69.9 | 83.5 | 71.8 | 82.4 | 70.9 | 4.42 | 4.23 |
| | Horiguchi *et al.* [14], 2018 | - | - | - | - | - | 78.1 | 61.7 | - | - | 77.8 | 60.8 | - | - | 2.18 | 1.96 |
| | Wen *et al.* [27], 2021 | 80.3 | 67.6 | 79.3 | 76.2 | 62.3 | 81.4 | 67.8 | 79.6 | 66.7 | 79.9 | 66.1 | 79.3 | 66.9 | 4.31 | 3.91 |
| | Saeed *et al.* [7], 2022 | 82.1 | 72.2 | 81.7 | 77.0 | 63.9 | 83.3 | 72.8 | 81.1 | 71.7 | 82.9 | 72.4 | 83.2 | 72.1 | 5.63 | 5.98 |
| | Chen *et al.* [22], 2023 | 83.0 | 69.1 | 82.1 | 78.0 | 64.6 | 84.1 | 73.7 | 81.9 | 72.5 | 83.3 | 72.9 | 83.6 | 72.8 | 6.11 | 6.38 |
| Unsup. | Nagrani *et al.* [13], 2018 | 78.5 | 61.1 | 77.2 | 74.9 | 58.8 | 68.1 | 67.1 | 68.3 | 67.5 | 68.9 | 67.4 | 68.5 | 63.9 | 3.55 | 3.55 |
| | Morgado *et al.* [20], 2022 | 78.2 | 65.2 | 78.4 | 75.1 | 62.9 | 78.3 | 64.7 | 77.8 | 64.2 | 77.6 | 64.7 | 77.2 | 64.1 | 3.69 | 3.55 |
| | Chen *et al.* [5], 2022 | 82.0 | 71.8 | 81.6 | 77.3 | 64.2 | 83.2 | 71.3 | 81.9 | 70.0 | 82.5 | 71.6 | 82.0 | 71.3 | 5.59 | 5.78 |
| | Zhu *et al.* [21], 2022 | 82.6 | 68.6 | 81.3 | 77.8 | 63.7 | 82.2 | 67.4 | 81.1 | 67.1 | 81.7 | 67.4 | 81.1 | 66.8 | 3.80 | 3.70 |
| | *Proposed* | **86.1** | **77.9** | **84.9** | **80.8** | **68.8** | **86.4** | **76.9** | **84.8** | **75.8** | **85.6** | **77.3** | **86.2** | **76.5** | **7.56** | **7.54** |

The verification results in Table 4 show that the feature subnetworks affect the overall performance but do not change the performance ranks with respect to other methods. In other words, the features extracted using the subnetworks in [7] decrease the overall performance of all methods compared to the results in Table 3. Though, the proposed method shows outstanding results relative to other methods despite performance degradation.

## 4.7 Evaluation on Other Dataset Splits

For cross-validation, we compared the results with state-of-the-art methods [5, 7, 13–15, 20–22, 27] using the different data set splitting configurations used in other recent studies [7, 13] for cross-modal verification, matching, and retrieval.

The results are summarized in Table 5. The results show significant performance gains of the proposed method over state-of-the-art methods, demonstrating the superiority of our approach. Furthermore, the results are analogous to those of Table 3, where the proposed method outperforms other methods in all cross-modal association tasks tested in our experiments, *i.e.* cross-modal verification, matching, and retrieval tasks.

## 4.8 Evaluation for Feature Convexity

The proposed method increases the convexity of the embedded features by helping the cross-modal features to be gathered. Thus, we conducted additional experiments to quantitatively verify improvements in feature embedding over previous methods.

We evaluated the effectiveness of our method by analyzing the distribution of embedded features for face, voice, and cross-modal inputs. Specifically, we measured both the intra-class and inter-class cosine distances across modalities to assess clustering quality and discriminative capability.

For intra-class evaluation, we computed the average cosine distance between all pairs of features (face-face, voice-voice, and face-voice) that belong to the same identity. This measures how closely features of the same individual are clustered in the embedding space. In contrast, inter-class evaluation was conducted by averaging the cosine distances between features belonging to different identities. This captures the degree of separation across different classes in the shared feature space. By jointly considering intra-class compactness and inter-class separability, we are able to more comprehensively evaluate the quality of the learned cross-modal embedding. The distance metric ($= 1 - \textit{cosine similarity}$) is used to measure distance for fair comparisons. Thus, the result values range from 0 to 1, with lower values indicating well-clustered features. The results are summarized in Table 1.

The proposed method shows significantly lower intra-class embedding distances compared to Saeed [7] and Nagrani [13], and achieves comparable or better results than Chen [5, 22]. In particular, our method achieves the lowest intra-class cross-modal distance (*i.e.*, 0.0053), indicating that voice and face features belonging to the same identity are effectively aligned in the shared embedding space. In addition to intra-class similarity, we also report inter-class distances to analyze the discriminative power of each method. As shown in the right columns of Table 1, the proposed method yields the highest inter-class distances across all modalities. This result demonstrates that the proposed embedding space not only tightly clusters features of the same identity, but also enforces stronger separation between different classes. In particular, the inter-class distance for cross-modal features (0.1987) is substantially larger than that of existing methods, showing improved class separability.

**Table 6**: Verification on AVSpeech Dataset. Despite the different domain and larger data scale, our proposed method continues to show superior AUC and lower EER, further confirming its adaptability and generalization across datasets.

| Method | Verification (U) | |
|---|---|---|
| | AUC % (↑) | EER % (↓) |
| Nagrani *et al.* [13], 2018 | 81.4 | 27.1 |
| Saeed *et al.* [7], 2022 | 84.9 | 25.5 |
| Chen *et al.* [5], 2022 | 85.3 | 25.2 |
| Chen *et al.* [22], 2023 | 86.1 | 23.2 |
| *Proposed* | **88.2** | **19.9** |

The observed results validate the effectiveness of our convex feature embedding strategy. By introducing intermediate features between modalities and optimizing them to lie within intra-class convex regions, the model learns to associate face and voice features more robustly while maintaining clear inter-class boundaries. This dual effect—of pulling together same-class features and pushing apart different-class ones—enhances both *modality alignment* and *identity discrimination*, which are critical for reliable face-voice association in unconstrained environments.

## 4.9 Evaluation on Other Face-Voice Dataset

For further evaluations, we performed cross-modal verification tasks on AVSpeech [89] dataset, a large-scale audio-visual dataset comprising speech video clips without interfering background noise composed of 290k YouTube videos. Without loss of generality, we compared our method with three recent methods used in Table 4. The results are summarized in 6.

The results are consistent with those of Tables 3, 4, and 5. Throughout the experiments, the proposed method shows the lowest AUC and the highest EER over the previous methods, regardless of feature subnetworks, data splits, datasets, and cross-modal tasks. The varied and consistent results demonstrate the effectiveness of our methods in the association of face and voice.

## 4.10 Retrieval Example

To qualitatively assess the effectiveness of the proposed method, we present visualizations of cross-modal retrieval results in Fig. 6. We include a comparative baseline, Chen *et al.* [22], which demonstrated competitive performance across multiple tasks in our quantitative benchmarks (see Table 3 and Table 5). Both face-to-voice and voice-to-face retrieval tasks are evaluated to provide a bidirectional analysis.

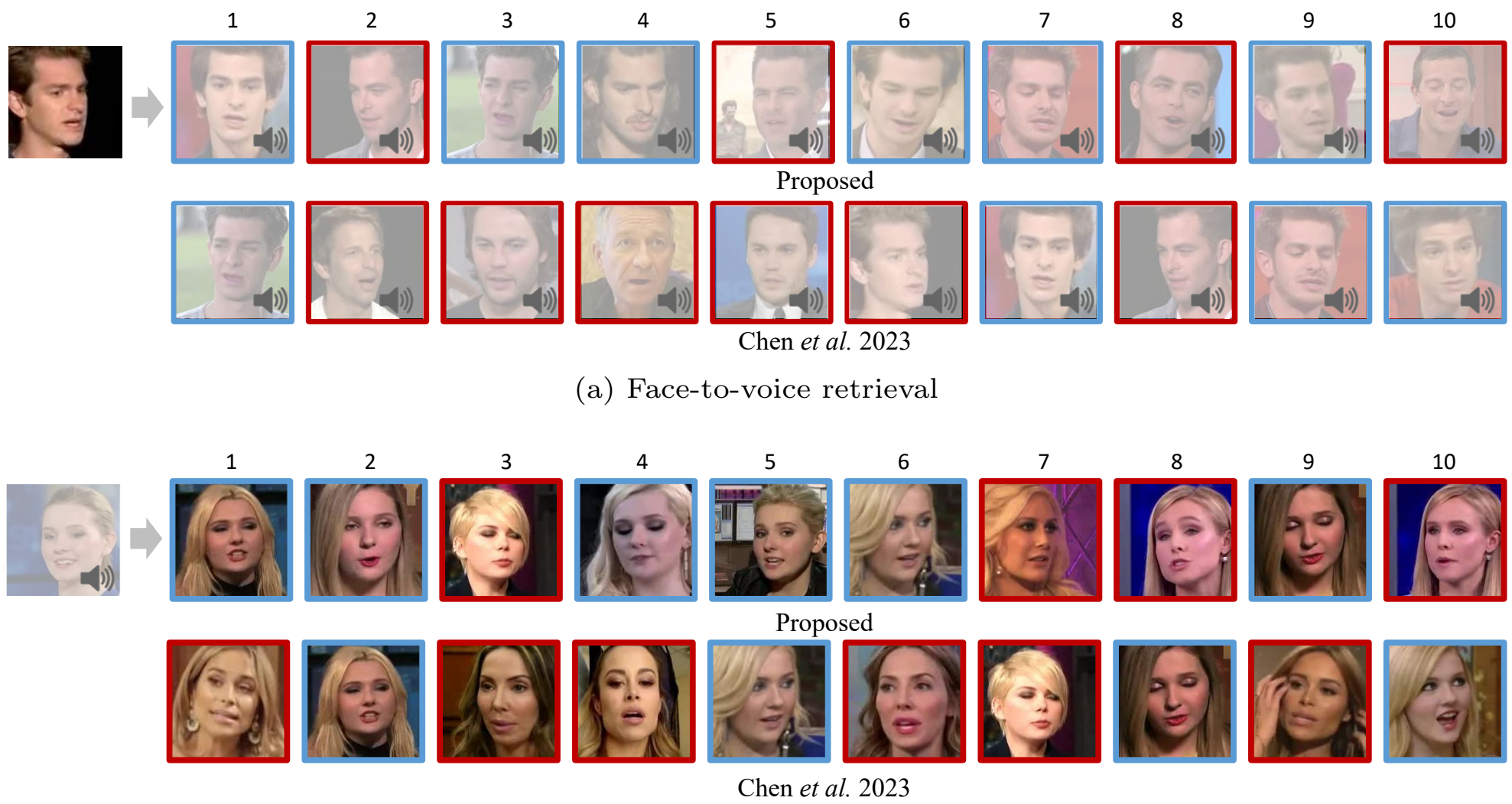

(a) Face-to-voice retrieval

(b) Voice-to-face retrieval

**Fig. 6**: Cross-modal retrieval examples, displayed in descending order of similarity. The proposed method appropriately matches (a) faces from voices and (b) voices from faces, respectively, indicating the commonality between these two modalities. In particular, correct matches are highlighted in blue and incorrect matches in red, demonstrating that our method effectively learns shared identity cues across face and voice embeddings.

In each example, a query from one modality (face or voice) is used to retrieve the top-10 most similar samples from the other modality based on cosine similarity in the shared embedding space. Retrieved samples are sorted in descending order of similarity. Identity-consistent (correct) retrievals are outlined in blue, and mismatched (incorrect) ones in red.

In the face-to-voice scenario (Fig. 6 (a)), the proposed method successfully retrieves 6 voice samples corresponding to the query face, while Chen *et al.*retrieves 4 correct matches. In the voice-to-face scenario (Fig. 6 (b)), our method returns 6 correct identities, compared to 4 by the baseline. In both directions, there is an approximate 50% overlap in the top retrieved samples, suggesting that both models leverage shared modality-invariant characteristics, such as prosody or visual identity traits.

Nevertheless, the proposed method demonstrates superior retrieval accuracy and consistency. This improvement is attributed to the convex feature embedding and cross-modal attention mechanisms, which jointly encourage intra-class compactness and inter-class separation across modalities. These mechanisms guide identity-specific face and voice features into a shared convex region, resulting in improved modality alignment and retrieval discriminability.

Even in failure cases, both methods often retrieve semantically plausible results, such as individuals with similar facial appearance or vocal tone. However, the proposed method more consistently places identity-correct samples at the top of the retrieval ranking, highlighting its improved capacity for identity preservation across modalities.

These qualitative results corroborate the quantitative findings reported in earlier sections and emphasize the benefit of explicitly modeling modality-invariant yet identity-specific representations for robust face-voice association.

### 4.11 Limitations and Future Work

While our proposed framework demonstrates robust performance in cross-modal verification, matching, and retrieval tasks, it is imperative to acknowledge inherent limitations within face-voice association systems and to propose avenues for future research.

Our methodology, akin to prior studies [5, 7, 13], operates under the premise that individuals with similar facial features are likely to possess analogous vocal characteristics, and vice versa. This hypothesis has been statistically substantiated by the elevated accuracy rates reported in preceding face-voice matching and retrieval experiments. Nonetheless, this assumption is probabilistic and does not universally apply to all individuals. Indeed, numerous instances exist where a person's voice and facial appearance do not exhibit such cross-modal resemblance.

These deviations from the presumed cross-modal correlation can precipitate failures in association tasks. For example, in Fig. 4(a), our proposed method generally aligns face and voice embeddings within a shared convex region. However, the cyan-colored feature cluster illustrates a failure case where, despite proximity, face and voice features remain segregated in the feature space. This observation suggests that certain individuals may lack typical identity-related cues across modalities, leading to modality mismatches even in well-trained systems.

Such cases underscore a fundamental limitation of face-voice association: the assumption of a shared latent identity space may not hold for individuals with highly divergent facial and vocal characteristics. To mitigate this issue, future systems could integrate richer paralinguistic voice features—such as prosody, speaking style, or articulation habits, which have been shown to convey identity-specific cues [90]. Incorporating these elements may reduce ambiguity in associations involving non-prototypical subjects.

Additionally, Fig. 4 reveals the presence of dispersed embeddings distant from their corresponding class clusters. These features, identifiable as outliers, likely result from pronounced intra-class variation or compromised signal quality. Given that datasets like VoxCeleb [4] are sourced from unconstrained online platforms such as YouTube, variations in age, environment, and recording quality are prevalent, leading to substantial intra-class variability.

It is noteworthy that all experiments were conducted using a unified, fixed dataset protocol to ensure fair comparisons with prior work [5, 22]. However, in practical applications, performance could be enhanced by implementing automatic quality estimation or anomaly detection methods to identify and exclude such outlier samples. Moreover, incorporating metadata, such as age or signal quality indicators (*e.g.*, signal-to-noise ratio for voice or resolution for face), into the embedding process could further account for these factors. Leveraging such auxiliary information presents a promising direction for enhancing robustness in real-world deployments.

In summary, we identify two primary challenges: (1) individuals who deviate from general face-voice similarity trends, and (2) outliers arising from noise, age variation, or suboptimal quality. Addressing these challenges may involve integrating cross-modal embeddings with higher-level semantic features or conditioning associations on supplementary metadata. We propose these directions for future exploration.

# 5 Conclusion

In this paper, we have proposed a simple yet powerful cross-modal feature learning approach for the voice-face association. The proposed method has introduced intermediate features between face and voice to embed cross-modal features in convex feature space by learning other features of the same label between cross-modal features. In addition, we have presented a cross-modal attention mechanism to efficiently fuse the feature embeddings. The convex embedding and the cross-modal attention mechanism have allowed cross-modal attention to significantly reduce inter-modal gaps and inter-class variances. In extensive experimental results on cross-modal verification, matching, and retrieval on the large-scale dataset, the proposed method has achieved a distinct association between faces and voices of the same identity and outperformed existing state-of-the-art methods. This demonstrates that the proposed method significantly helps a network learn joint-modal features for cross-modal association. We hope that our results and discussion would be helpful in other cross-modal association work, such as cross-modal generation and separation.

# Declarations

## Author contribution

Taewan Kim: Software, Validation, Writing−original draft preparation, Jiwoo Kang: Conceptualization, Methodology, Investigation, Writing−review and editing, Supervision.

## 5.1 Funding

This work was supported by Institute of Information & communications Technology Planning & Evaluation (IITP) grant funded by the Korea government (MSIT) (No. RS-2023-00229451, Interoperable Digital Human (Avatar) Interlocking Technology Between Heterogeneous Platforms).

## Data availability

We used publicly accessible datasets online: Voxceleb in https://www.robots.ox.ac.uk/~vgg/data/voxceleb/ and AVSpeech in https://looking-to-listen.github.io/avspeech/.

## Conflict of interest

The authors declare no competing interests.

## Ethics approval

Not applicable.